\documentclass{article}

\usepackage{microtype}
\usepackage{graphicx}
\usepackage{subfigure}
\usepackage{booktabs} 

\usepackage{hyperref}

\usepackage[accepted]{icml2025}

\usepackage{amsmath}
\usepackage{amssymb}
\usepackage{mathtools}
\usepackage{amsthm}

\usepackage[capitalize,noabbrev]{cleveref}

\theoremstyle{plain}

\theoremstyle{definition}

\theoremstyle{remark}

\usepackage[textsize=tiny]{todonotes}

\icmltitlerunning{NGENT: Next-Generation AI Agents Must Integrate Multi-Domain Abilities to Achieve Artificial General Intelligence}

\begin{document}

\twocolumn[
\icmltitle{NGENT: Next-Generation AI Agents Must Integrate Multi-Domain Abilities to Achieve Artificial General Intelligence}



\icmlsetsymbol{equal}{*}

\begin{icmlauthorlist}
\icmlauthor{Zhicong Li}{rmu}
\icmlauthor{Hangyu Mao}{pku}
\icmlauthor{Jiangjin Yin}{hzau}
\icmlauthor{Mingzhe Xing}{pku}
\icmlauthor{Zhiwei Xu}{sdu}
\icmlauthor{Yuanxing Zhang}{pku}
\icmlauthor{Yang Xiao}{pku}
\end{icmlauthorlist}

\icmlaffiliation{rmu}{Renmin University, China}
\icmlaffiliation{pku}{Peking University, China}
\icmlaffiliation{hzau}{College of Informatics, Huazhong Agricultural University, China}
\icmlaffiliation{sdu}{Shandong University, China}

\icmlcorrespondingauthor{Hangyu Mao}{hy.mao@pku.edu.cn}

\icmlkeywords{Machine Learning, AI Agent, Next-Generation AI Agent, LLM-based Agent}

\vskip 0.3in
]



\printAffiliationsAndNotice{}  

\begin{abstract}
This paper argues that the \underline{n}ext \underline{g}eneration of AI ag\underline{ent} (NGENT) should integrate across-domain abilities to advance toward Artificial General Intelligence (AGI). Although current AI agents are effective in specialized tasks such as robotics, role-playing, and tool-using, they remain confined to narrow domains. We propose that future AI agents should synthesize the strengths of these specialized systems into a unified framework capable of operating across text, vision, robotics, reinforcement learning, emotional intelligence, and beyond. This integration is not only feasible but also essential for achieving the versatility and adaptability that characterize human intelligence. The convergence of technologies across AI domains, coupled with increasing user demand for cross-domain capabilities, suggests that such integration is within reach. Ultimately, the development of these versatile agents is a critical step toward realizing AGI. This paper explores the rationale for this shift, potential pathways for achieving it. 
\end{abstract}

\section{Introduction}
Broadly speaking, AI agents are systems designed to autonomously perform tasks (take actions) in pursuit of specific objectives, using information gathered from their environment. The key components of these agents include: Perception—the process of gathering information about the environment (e.g., through sensors or data inputs); Reasoning and Planning—the ability to interpret this information and make decisions; Action—executing these decisions through physical or virtual means (e.g., moving a robot or sending a message); and Learning—the capability to adapt behavior based on feedback or new data.

The evolution of AI agents can be broadly divided into three stages. (1) Rule-based AI Agents: Early agents were simple, rule-based systems designed to perform narrowly defined tasks \cite{franklin1997autonomous, castelfranchi1998modelling, alonso2002ai}. Notable examples include systems like Deep Blue for chess \cite{campbell2002deep} and early expert systems used for specialized functions, such as medical diagnosis \cite{shortliffe1986medical}. (2) Machine Learning-based AI Agents: In contrast, modern agents leverage machine learning \cite{mitchell1997machine}, particularly deep learning \cite{lecun2015deep}, allowing them to handle more complex tasks such as dialog generation \cite{li2016deep}, image recognition \cite{krizhevsky2012imagenet}, and autonomous navigation \cite{bagnell2010learning}. These agents demonstrate a much greater degree of adaptability and flexibility than their predecessors. Despite these advancements, both rule-based and learning-based agents typically excel within specific domains and face challenges when attempting to transfer knowledge or skills across different tasks. (3) Foundation Model-based AI Agents: Both Large Language Model (LLM)-based and Large Multimodal Model (LMM)-based agents have brought profound changes to the agent landscape \cite{xi2023rise, wang2024survey}. Equipped with the vast knowledge embedded in LLMs and LMMs, these AI agents are capable of performing a \emph{much broader} array of tasks compared to their predecessors, ranging from role-playing \cite{shao2023character,chen2024persona}, tool usage \cite{schick2023toolformer,ruan2023tptu} and coding development \cite{roziere2023code,li2024pet,zhang2024pybench} to computer control \cite{hong2024cogagent,niu2024screenagent}.

\begin{figure*}[th!]
\vskip 0.2in
\begin{center}
\centerline{\includegraphics[width=1.0\linewidth]{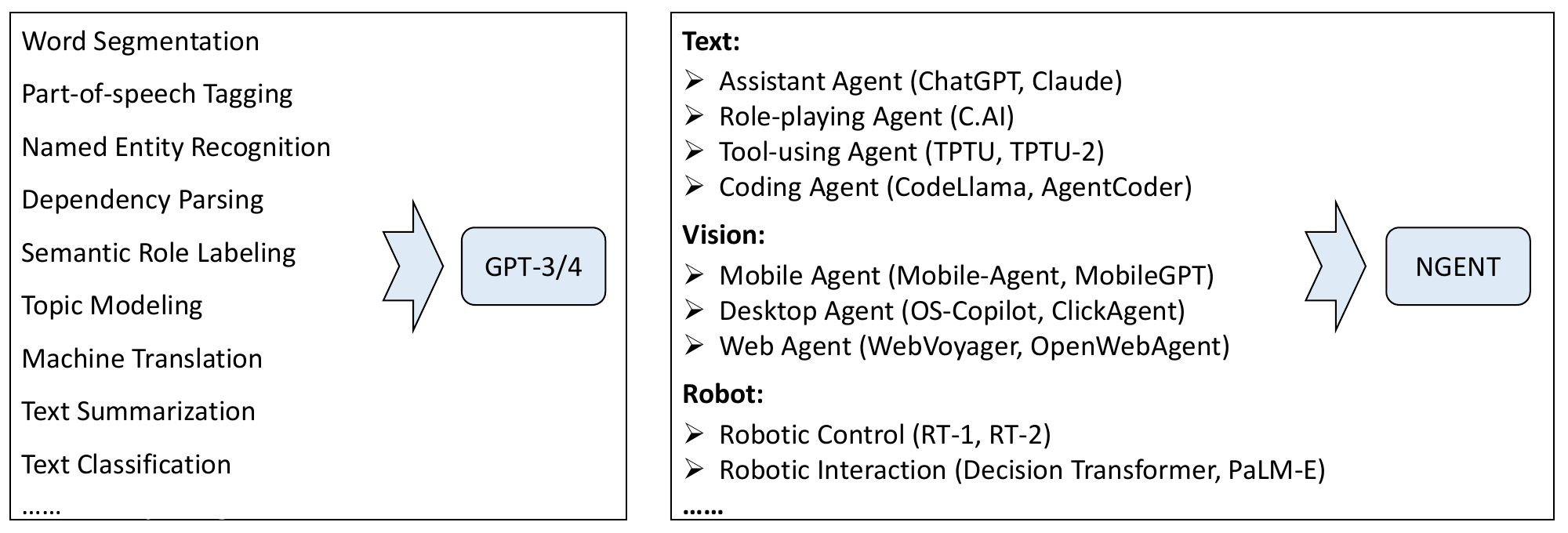}}
\caption{The position of this paper. Similar to how GPT-3 and GPT-4 revolutionized natural language processing by excelling in virtually all NLP tasks, we propose that the next generation of AI agents should demonstrate comparable versatility across a wide range of domains.}
\label{figure:motivation}
\end{center}
\vskip -0.2in
\end{figure*}

Despite the broad capabilities of these AI agents, they still remain limited by their specific domains. For example, while role-playing agents can convincingly portray different personalities with nuanced tones, and tool-using agents can leverage various tools like weather and map APIs, their coding and computer control abilities often remain subpar. The reverse is also true: agents excelling in coding or computer control may struggle with tasks in other domains. This limitation holds true even for agents powered by top-tier LLMs such as OpenAI O1 \cite{jaech2024openai}. 

We believe that the next leap in AI development must focus on overcoming the limitations of domain-specific models. As demonstrated in Figure \ref{figure:motivation}, \emph{in much the same way that GPT-3 \cite{brown2020language} and GPT-4 \cite{achiam2023gpt} revolutionized natural language processing (NLP) by demonstrating proficiency across virtually all NLP tasks, we argue that the next generation of AI agents should exhibit a similar versatility across multiple domains}, from personal assistants and role-playing to coding, tool usage, mathematical problem-solving, and more. This holistic approach would mark a significant step towards achieving AGI.

Achieving this vision will require overcoming several key challenges, including the integration of different specialized capabilities into a unified architecture, the development of multi-task learning models, and the ability to maintain high performance across varied use cases. We argue that a shift toward these next-generation agents is not only feasible, but crucial for the continued progress of the AI field. Throughout this paper, we outline the rationale for this shift and explore potential pathways to realize it. 

\section{Support}
We begin by discussing the overarching goal of AGI and the growing user demand, which underscore the \emph{necessity} for next-generation AI agents to integrate across domains. We then present the convergence of techniques, highlighting the \emph{feasibility} of such integration in future AI agents.

\subsection{Towards AGI Research}
Cognitive science suggests that true intelligence is not narrowly specialized, but flexible and adaptive across multiple domains \cite{norman1980twelve,sternberg1999theory}. Humans, for example, do not excel in just one task; they continuously switch between roles, contexts, and challenges.

The ultimate goal of AGI is the creation of a machine that can understand, learn, and apply knowledge across a wide variety of tasks in a manner comparable to human intelligence. Existing AI systems (and specifically the AI agents discussed in this paper), though impressive, are still far from achieving this goal. The development of next-generation AI agents that integrate diverse capabilities is a crucial step on the path toward AGI.

\subsection{User Demand for Versatile AI Systems}
The increasing demand for versatile, all-encompassing AI systems is driven by both consumer and enterprise needs. Current personal assistants, such as Google Assistant and Amazon's Alexa, provide a broad range of services, but their capabilities are still limited by a lack of integration. A user who needs a tool that can automate repetitive tasks in the workplace, while also providing technical assistance and offering personalized advice, must often rely on several different systems.

Next-generation agents could combine these functionalities into a single interface, streamlining user interactions. This would offer tangible benefits in terms of time savings, efficiency, and the ability to tackle multi-step workflows in a cohesive manner. For example, a general-purpose AI agent could automate technical troubleshooting, provide guidance on creative endeavors, and assist with personal management tasks, all within the same context.

\subsection{Technological Convergence}
\subsubsection{Model Architecture Convergence}
\textbf{For LLM-based Agents.} It is evident that Transformer-based architectures form the core foundation of GPT-4, which has demonstrated exceptional performance across a broad spectrum of NLP tasks. 

Building on model architectures like GPT-4, researchers have developed various types of LLM-based AI agents: some focus on highly intelligent assistants, such as ChatGPT, which excel in problem-solving and information retrieval across diverse domains; others emphasize personified role-playing agents, like Character.AI, which specialize in simulating human-like personalities for engaging interactions. Additionally, tool-using agents for automating workflows \cite{qu2024tool, ruan2023tptu, kong2024tptu, kong2023tptuarxiv} and coding assistants for software development \cite{roziere2023code, guo2024deepseek, li2024pet, zhang2024pybench} are also widely recognized.

\textbf{For LMM-based Agents.} Current AI systems are rapidly advancing in their ability to process multiple types of inputs simultaneously. For instance, OpenAI’s GPT-4o \cite{hurst2024gpt} represents a significant step towards more natural human-computer interaction, as it can accept any combination of text, audio, image, and video as input, and generate any combination of text, audio, and image outputs. This ``One-For-All'' model is also based on the Transformer architecture, with different tokenizers designed for different modalities of data. Research is also increasingly focused on developing unified Transformer models capable of handling multimodal tasks effectively \cite{wang2022ofa}.

For multimodal AI agents, a key area of focus is operating system (OS) agents, which aim to control computing devices through mobile apps \cite{wang2024mobile,lee2024mobilegpt,zhang2023appagent}, desktop applications \cite{wu2024copilot,hoscilowicz2024clickagent}, or web interfaces \cite{he2024webvoyager,iong2024openwebagent,koh2024visualwebarena}, thereby significantly enhancing the efficiency of user-device interactions. Additionally, there are other LMM-based agents, such as Llava-plus \cite{liu2025llava} for vision tool-using agents, as well as a detailed survey \cite{xie2024large}. The model architecture of these agents is also primarily based on the ``One-For-All'' Transformer. 

\textbf{For Robotic Agents and Beyond.} We observe a similar trend across other domains, where there is an increasing reliance on Transformer-based models. For example, in the reinforcement learning (RL) field, the Decision Transformer \cite{chen2021decision} and Trajectory Transformer \cite{janner2021offline} have become dominant architectures for offline RL agents, while the original Transformer has been directly used or other Transformer architectures have been developed for online RL agents \cite{zheng2022online,mao2022transformer,mao2023pdit,zhangsequential}. In the robotics domain, architectures like RT-1 \cite{brohan2022rt} and RT-2 \cite{brohan2023rt} are designed with minor modifications to the original Transformer architecture.

\textbf{Summarization.} Across various AI agents—whether LLM, LMM, robotics, or RL—there is a clear trend toward the adoption and adaptation of Transformer-based architectures. This convergence suggests the potential for the next generation of AI agents to be integrated into a unified system capable of handling cross-domain tasks, including those related to text, vision, robotics, and more.

\subsubsection{Advancement in Learning Algorithms}
It is clear that advances in multi-task learning, transfer learning, self-supervised learning, meta-learning, and reinforcement learning \footnote{We will not introduce the concepts of these algorithms, as their role in enabling next-generation AI agents is self-evident.} are converging to provide the necessary foundation for next-generation AI agents. These algorithmic innovations empower agents to learn from and adapt to a wide variety of domains, leveraging abundant unlabeled data and self-improvement capabilities, ultimately facilitating the integration of specialized capabilities into a unified, versatile system.

Recently, tool learning \cite{qin2023tool} has gained significant attention as a powerful technique. By equipping AI agents with various tools, they can tackle diverse tasks across different domains. For example, HuggingGPT \cite{shen2024hugginggpt} leverages a range of AI models from Hugging Face to address a broad spectrum of sophisticated AI-related tasks, spanning language, vision, speech, and more, achieving impressive results across these modalities. Similarly, ToolLLM \cite{qin2023toolllm} masters over 16,000 real-world APIs to solve daily tasks, extending its capabilities beyond AI-specific problems (e.g., stock trading, ordering food, and more). Many recent tool-learning methods follow this approach, such as Toolformer \cite{schick2023toolformer}, TPTU \cite{ruan2023tptu}, TPTU-2 \cite{kong2024tptu}, and Chameleon \cite{lu2024chameleon}, further demonstrating the potential of integrating tools to enhance the versatility and capabilities of next-generation AI agents.

\textbf{Summarization.} It seems that combining tool learning with multi-task learning could lead to the development of truly next-generation AI agents (NGENTs)—that is, a single AI agent capable of handling all tasks, given an expansive set of tools that covers the full spectrum of task types.

\subsubsection{Convergence of Components in AI Agents}
The convergence of components in AI agents is becoming increasingly evident. As outlined in the first section, the core components of AI agents include Perception, Reasoning and Planning, Action, and Learning. Perception involves gathering environmental data (e.g., through sensors or data inputs), while Reasoning and Planning enable agents to interpret this information and make decisions. Action is the execution of these decisions, either physically (e.g., moving a robot) or virtually (e.g., sending a message). Learning allows AI agents to adapt their behavior based on feedback or new data. In addition, some AI agents integrate other components, such as short- and long-term memory, which help track historical context, improve decision-making, and reduce redundant actions.

Underlying these components are various techniques we have discussed, such as using different tools for Perception and Action, leveraging Transformer-based foundation models for Reasoning and Planning, and employing diverse algorithms for Learning. 

\textbf{Summarization.} The development of modular, scalable AI agents enables components to be easily swapped or replaced based on the task at hand, providing substantial flexibility. This modularity is essential for creating unified AI agents capable of efficiently performing a wide range of tasks across diverse domains.

\subsection{Summarization of Support}
The ultimate goal of AGI is to develop AI that can perform a broad range of tasks, akin to human intelligence. The growing demand for versatile systems requires AI agents capable of integrating multiple functionalities into a single interface. Technological convergence is driving this progress: Transformer-based models are increasingly utilized across various AI domains, while advancements in learning algorithms, such as multi-task and tool learning, enable agents to handle diverse tasks. Furthermore, modular AI components enhance flexibility and adaptability. These trends are paving the way for next-generation AI agents that can autonomously learn, adapt, and execute tasks across different domains. We advocate that researchers should focus more on this integrated approach, advancing the development of next-generation AI agents, while placing less emphasis on continuing fragmented research on first-generation AI agents that address isolated domains.

\section{A Possible Way to NGENT}
\begin{figure*}[th]
\vskip 0.2in
\begin{center}
\centerline{\includegraphics[width=0.88\columnwidth]{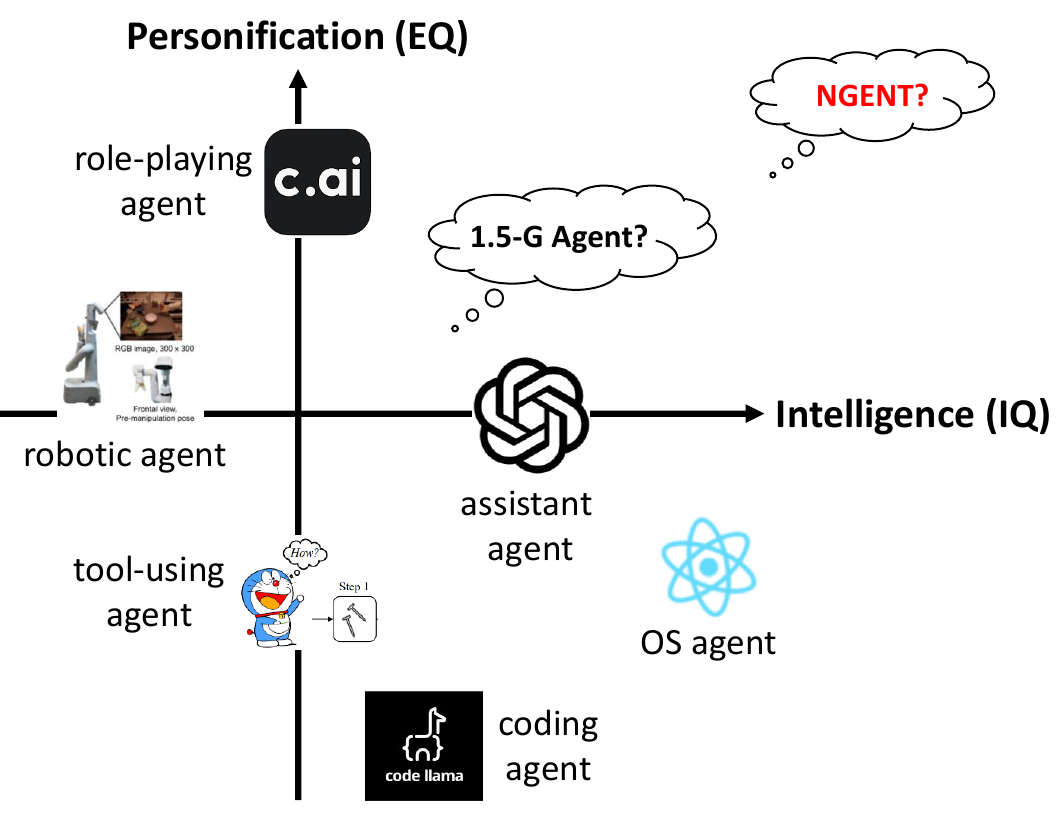}}
\caption{The landscape of foundation model-based AI agents. Current research primarily concentrates on specific domains, such as intelligent assistants (e.g., ChatGPT), role-playing (e.g., C.AI), coding (e.g., CodeLlama \cite{roziere2023code} and Deepseek-Coder \cite{guo2024deepseek}), tool usage (e.g., TPTU \cite{ruan2023tptu} and TPTU-2 \cite{kong2024tptu}), OS operations (e.g., OS-copilot \cite{wu2024copilot}), or robotic control (e.g., RT-1 \cite{brohan2022rt} and RT-2 \cite{brohan2023rt}). However, we believe that the next generation agents must go beyond individual specializations and instead embody general capabilities aimed at achieving AGI, seamlessly integrating diverse skills to operate across a wide range of tasks and contexts. In this paper, we propose a step towards this goal by introducing the concept of a ``1.5-generation agent'', which focuses on the critical challenge of integrating intelligence with personification, ultimately paving the way for the development of the next-generation AI agent (NGENT).}
\label{figure:path}
\end{center}
\vskip -0.2in
\end{figure*}

Given the support discussed above, particularly the convergence of various techniques, the design and realization of next-generation AI agents appear both feasible and imminent. In fact, we believe that integrating these capabilities into a single AI agent is a near-term possibility.

However, a key challenge remains: different tasks exhibit distinct distributions. Most of the tasks discussed earlier tend to be IQ-based, but there exists a separate class of tasks that focus on emotional intelligence (EQ). Integrating both IQ and EQ-based tasks into a single AI agent becomes more challenging, as these tasks often pull in different directions. An AI agent with high IQ typically requires a focus on abstract reasoning, efficiency, and factual accuracy, whereas an AI agent with high EQ demands a more nuanced approach, emphasizing emotional intelligence, empathy, and engaging communication.

As illustrated in Figure \ref{figure:path}, a promising approach is to first achieve a balance between IQ and EQ by integrating intelligent assistance with a user-friendly persona. While these two paradigms may seem distinct, they are not entirely independent. In fact, the intelligence required for effective persona management can be seen as an extension of the capabilities found in intelligent assistants, particularly in dialogue, task management, and personalized interactions. 

This insight points to the potential for a ``1.5-generation agent'' (1.5-G Agent). By initially developing a hybrid model that successfully integrates intelligent assistance with a user-friendly persona, we can lay the groundwork for future evolution. In this model, the intelligent assistant would take precedence in core functional tasks such as coding, mathematical problem-solving, and tool usage, while the persona aspect would be optimized for engaging, human-like interaction. Rather than being a separate feature, the persona would be deeply integrated with the agent's core intelligence, enhancing the user experience without compromising its functional capabilities.

The ``1.5-generation agent'' model could serve as a stepping stone toward the full 2.0 version of the next-generation agent (2.0-G NGENT), one capable of performing a wide range of both IQ and EQ-based tasks, without requiring specialization in any particular domain.

Ultimately, while the journey to a fully realized AGI system may still be distant, the 2.0-G NGENT will provide valuable insights into achieving this goal.

\section{Preliminary Experiments}
In this section, we investigate whether the ``1.5-generation agent'' is possible. We have developed a comprehensive training pipeline consisting of three stages, Instruction Pre-training (IPT), Supervised Fine-Tuning (SFT) and DPO Training. Each stage is designed to address different aspects of the agent's capabilities, building on the previous one to gradually refine the model's balance between IQ and EQ.

\subsection{Instruction Pre-Training}
The traditional Continue Pre-Training (CPT) method involves further training and optimizing an existing LLM for specific application domains. This approach typically entails higher computational resource expenditure and increased time costs. Therefore, we propose a novel IPT method to replace CPT, which achieves excellent experimental results using only approximately 1\% of the training data.

Specifically, in our work, we focus on the agent's personification capabilities and intellectual level during multi-turn conversations with users. Given that most base models inherently possess a high level of intelligence but lack personification, we emphasize the introduction of personalized datasets during the IPT phase to address this deficiency.
In detail, we collect dialogues from novels, scripts, and social media as our dataset. This type of data is naturally dense with extensive character settings and real-life conversations, using which can enhance the agent's ability to personalize. 

Formally, consider the concatenation $\mathcal{U}=\{\mathcal N, \mathcal C, \mathcal D\}$, where $\mathcal N$ describes the background narration, $\mathcal C$ represents the character name, and $\mathcal D$ refers to the dialogue. Given the tokens before turn $t$, the IPT objective over a dataset $\mathcal{U}$ can then be formulated as follows:
\begin{equation}\label{ipt_loss}
\begin{aligned}
    \mathcal L_{IPT}&=-\log p(\mathcal{D}|\mathcal{N},\mathcal{C})\\
    & =-\sum_{t=1}^{T_{\mathcal{D}}} \log p(\mathcal{D}_{t}|\mathcal{D}_{<t},\mathcal{N}_{<t},\mathcal{C}_{<t}),
\end{aligned}
\end{equation}

where $T_{\mathcal{D}}$ is the total dialog turn of $\mathcal U$.  In other words, we compute the loss associated with the character dialogue portions, while retaining character names and background narration. This approach aids the model in understanding each character's traits, thereby effectively learning different personalities.

\subsection{Supervised Fine-Tuning}
In this stage, the primary goal is to activate and enhance the agent's ability to simulate personality, without diminishing or even enhancing its intelligence. To achieve this, we propose an Personified Style Rewriter (PSR) and develop a comprehensive high-quality data generation process that relies on minimal manually annotated seed data. Additionally, an iterative training paradigm is introduced to significantly boost the agent's performance on both  personification and intelligence.

\subsubsection{Data Preparation}\label{data_preparation}
\paragraph{Synthetic Personality Data Generation.}
To prevent agent overfitting into single personality, it is essential to creat a vast range of personality role cards to enhance the agent's ability to understand and adapt to various personalities.
We collect over 100K personality roles from various sources, including historical figures, film IPs, virtual characters, etc. The personality traits are then extracted based on the basic information, characteristics, and dialogue styles of each character, thereby constructing personality profiles corresponding to different characters.

Upon acquiring an extensive collection of personality role cards, we utilized these along with a pre-prepared seed question dataset to generate and manually refine a series of high-quality seed data by emulating these personalities. Following this, we employ both the profiles and the corresponding seed data examples as few-shots to generate a large volume of responses.

While these responses encapsulate the personality traits delineated in the profiles, replies from some modern character role cards are still excessively formal. 
To address this, we propose the Personified  Style Rewriter (PSR) to transform these replies into more anthropomorphized responses.

\paragraph{PSR Model Training and Response Enhancement.} Given their role as assistants, most existing LLMs typically maintain a formal tone, even when instructed to simulate human-like responses. This often makes it apparent to users that they are interacting with a robot, not a human. Therefore, personifying the style of responses is crucial for enhancing personality attributes. To train a rewriter capable of such anthropomorphization, it is necessary to construct a specific dataset. However, the inherent formal tone of LLMs poses a challenge as it impedes the synthetic generation of anthropomorphized labels directly from existed powerful LLMs, like GPT-4. 

To overcome this issue, we adopt a contrary strategy by utilizing the strengths of LLM in generating formal responses. We prompt GPT-4 to rewrite \emph{a real user chat}, denoted as $c$. The rewritten result, $c'$, naturally exhibits a formal tone. These pairs $(c',c)$ are used as the training dataset to develop our PSR model.
Following this, we use the PSR to further refine the response from modern character role personality, resulting in the final synthesized personality dataset $D_{p}$.

\paragraph{Personality-Oriented Intelligent Contrastive Dataset Construction.} Training agents solely with the personality dataset $D_p$ leads to a significant problem: agents assigned specific personalities might avoid answering or pretend ignorance on queries it is inherently capable of resolving, causing a perceived decline in intelligence.  
This is because the personality-focused training data centers on persona dialogues, which contain fewer factual and informative elements.
To address this, we propose a novel personality-oriented intelligent contrastive dataset construction strategy to enhance the cognitive proficiency of the agent in a persona-specific context.

Specifically, for each data pair $(q_i,a_i)$ from the original SFT dataset, where $q_i$ denotes the query and $a_i$ the answer, we regenerate a tailored response $a^p_i$ reflective of each personality $p_i$. Subsequently, we incorporate both $(q_i, a_i)$ and $(q_i, a^p_i)$ into our dataset, thereby constructing a personality-focused intelligent contrastive dataset, denoted as $D_c$, which, combined with 
$D_p$, serves as our comprehensive final dataset $D = D_p \cup D_c$. Consequently, during the training phase, the agent not only preserves or augments its cognitive capabilities but also acquires the proficiency to respond appropriately across diverse personality profiles. 

\subsubsection{Iterative Supervised Fine-Tuning}
In traditional multi-turn instruction tuning, a training sample typically consists of $k$ query-response pairs, denoted as
\begin{equation}
    x=(q_1,r_1,q_2,r_2, ..., q_k,r_k),
\end{equation}
where $q_i$ and $r_i$ refers to query and response in $i$-th turn, respectively.
The model is trained by predicting the tokens in the response part, which is formulated in Equation \ref{response_loss}.
\begin{equation}\label{response_loss}
    \mathcal L_{response}=-\sum_{i=1}^L \log p(x_{i}|x_{<i}),x_i\in r.
\end{equation}
Here, $L$ refers to the total length of $x$ and $r$ signifies all the response segments within $x$.

Considering that human queries are often complex and diverse, it is challenging to comprehensively cover them through data collection alone. To this end, we introduce both an on-the-fly data augmentation method and an iterative SFT training paradigm.

Specifically, we first fine-tune the base model $M_0$ to develop an ask agent, $M_{ask}$, which simulates the user's role. This training method adheres to traditional SFT practices but incorporates a minor adjustment to the loss function such that only the tokens within the query segments are considered, as detailed in Equation (\ref{ask_loss}). 
Concurrently, on dataset $D$, we fine-tune $M_0$ according to Equation (\ref{response_loss}) to obtain $M_1$. 
This agent, $M_1$, interacts in real-time with $M_{ask}$ to generate new single and multi-turn dialogue data. This newly generated data is then filtered and combined with the original dataset to retrain our agent $M_0$ in an iterative cycle. The Algorithm \ref{alg:iter_sft} presents this process.

\begin{equation}\label{ask_loss}
    \mathcal L_{ask}=-\sum_{i=1}^L \log p(x_{i}|x_{<i}),x_i\in q.
\end{equation}

\begin{algorithm}[htbp]
\caption{Iterative Supervised Fine-tuning}\label{alg:iter_sft}
\begin{algorithmic}
\STATE {\bfseries Input:} Dataset $D$, base agent $M_0$, iteration number $t$
\STATE $M_1 \gets M_0$  trained on $D$ by Equation (\ref{response_loss})
\STATE $M_{ask} \gets M_0$ trained on $D$ by Equation (\ref{ask_loss})
\FOR{$i \gets 2 \text{ to } t$}
    \STATE $D_{i}$  $\gets$ interactions between $M_{i - 1}$ and $M_{ask}$
    \STATE $D\gets D \cup D_{i}$
    \STATE $M_i \gets M_0$ trained on $D$ by Equation (\ref{response_loss})
\ENDFOR
\STATE {\bfseries Output:} Chat agent $M_t$
\end{algorithmic}
\end{algorithm}


\subsection{Direct Preference Optimization}
To further enhance the quality of the agent's responses, we additionally incorporated a DPO training stage, This stage is designed to balance accuracy, engagement, and conciseness, and can also be tailored to meet specific product objectives. 

Specifically, we have collected 10K generated multi-turn data from SFT stage and construct
corresponding negative responses based on established criteria. These are used alongside positive responses to train the agent through the DPO algorithm.
Through this approach, the agent's anthropomorphic qualities and personality attributes are substantially improved, while simultaneously achieving greater engagement and conciseness in responses.

{
\begin{table*}[!th]
\caption{Main result on personification evalution. The best results are highlighted in \textbf{bold}, while suboptimal are \underline{underlined}.}
\label{eqresult}
\begin{center}
{
\footnotesize
\begin{tabular}{l|cccc|ccccc|ccc|c}
\midrule
& \multicolumn{4}{c|}{\textbf{Conversational Ability}}
& \multicolumn{5}{c|}{\textbf{Attractiveness}}              
& \multicolumn{3}{c|}{\textbf{Persona Consistency}}
& Overall \\ 
\cline{2-13}
& Flu. & Coh. & Cons. & \multicolumn{1}{c|}{Avg.}
& HL & CS & ED & Emp. & Avg.
& PB & PU & Avg.\\ \hline
ChatGLM3-6B & 3.269 & 3.647 & 3.283   &  \multicolumn{1}{c|}{3.399}  &   3.064 & 2.932 & 1.969 & 2.993 & 2.739 & 2.455 &    2.812 &    2.633  &   2.935        \\
XVERSE-7B  &   3.393 & 3.752 & 3.518   &  \multicolumn{1}{c|}{3.554}  &   3.395 & 2.743 & 2.013 & 2.936 & 2.772  &  2.564  &    2.887 &    2.726 &   3.022        \\
Baichuan2-7B   &  \multicolumn{1}{l}{3.551} & \multicolumn{1}{l}{3.894} & \multicolumn{1}{l}{3.827} & \multicolumn{1}{l|}{3.757}  &   \multicolumn{1}{l}{3.670} & \multicolumn{1}{l}{2.728} & \multicolumn{1}{l}{2.115} & \multicolumn{1}{l}{2.984} & \multicolumn{1}{l|}{2.874}  &    2.830    &   3.081 &    2.956    &  3.187                  \\
Qwen-7B    &   \multicolumn{1}{l}{3.187} & \multicolumn{1}{l}{3.564} & \multicolumn{1}{l}{3.229}   &  \multicolumn{1}{l|}{3.327}  &   \multicolumn{1}{l}{3.036} & \multicolumn{1}{l}{2.791} & \multicolumn{1}{l}{2.052} & \multicolumn{1}{l}{2.838} & \multicolumn{1}{l|}{2.679} &    2.605 &    2.780 &    2.693 &  2.898\\
InternLM-7B    &  3.527 & 3.823 & 3.744   &  \multicolumn{1}{c|}{3.698}  &   3.546 & 2.622 & 2.070 & 2.897 & 2.784   & 2.719 &    3.016 &    2.867    & 3.107                \\
XVERSE-13B &   3.444 & 3.811 & 3.559   &  \multicolumn{1}{c|}{3.605}  &   3.319 & 2.939       &   2.045 & 3.018 & 2.830   &    2.579 &    2.915 &    2.747  &   3.070    \\
Baichuan2-13B  &  3.596       &   3.924 & 3.864   &  3.795    & 3.700 & 2.703 & 2.136 & 3.021       &   2.890    &   2.808 &    3.081 &    2.944  &  3.204                \\
Qwen-14B   &   3.351 & 3.765 & 3.510   &  \multicolumn{1}{c|}{3.542}  &   3.354 & 2.871 & 2.237       &   2.970 & 2.858 &  2.744 &    2.900 &    2.822  &   3.078  \\
InternLM-20B   &  3.576 & 3.943    &   3.717   &  \multicolumn{1}{c|}{3.745}  &   3.582 & 2.885 & 2.132 & 3.047    &   2.911  &    2.753 &    3.041 &    2.897    &    3.186              \\
CharacterGLM   &  3.414 & 3.717 & 3.737   &  \multicolumn{1}{c|}{3.623}  &   3.738 & 2.265 & 1.966 & 2.812 & 2.695  & 2.301 &    2.969 &    2.635 &  2.991 \\
Xingchen   &   3.378 & 3.807 & 3.754   &  \multicolumn{1}{c|}{3.646}  &   3.757 & 2.272 & 2.100 & 2.799 & 2.732  &  2.772 &    3.055 &    2.913  & 3.077       \\
MiniMax    &   3.609    &   3.932       &   3.811   &  \multicolumn{1}{c|}{3.784}  &   3.768       &   2.672 & 2.150 & 3.017 & 2.902 &    2.774 &    3.125    &   2.950     &    3.207 \\
BC-NPC-Turbo & 3.578 & 3.898 & 3.916 &    3.798 &{\textbf{3.836}}    &   2.643 & 2.336    &   2.971 & 2.946 &   2.910 &   3.151 &   3.031  &   3.249 \\
GPT-3.5    &   2.629 & 2.917 & 2.700   &  \multicolumn{1}{c|}{2.749}  &   2.565 & 2.422 & 1.660 & 2.526 & 2.293   &   1.921 &    2.316 &    2.119   & 2.740     \\
GPT-4      &   3.332 & 3.669 & 3.343   &  \multicolumn{1}{c|}{3.448}  &   3.143 & 3.184    &   2.153 & 3.010 & 2.873    &2.721    &  2.873  &  2.797  &  3.048  \\

\textbf{NGENT (ours)} &{\textbf{3.794}} &{\textbf{4.114}} &{\textbf{3.923}} &{\textbf{3.944}} & 3.736 &{\textbf{3.453}} &{\textbf{3.100}} &{\textbf{3.361}} &{\textbf{3.413}} &{\textbf{3.518}} &{\textbf{3.307}} &{\textbf{3.113}} & {\textbf{3.523}} \\
\bottomrule             
\end{tabular}}
\end{center}
\end{table*}
}

\begin{table}[t]
\caption{IQ evaluation results.}
\label{iqresult}
\vskip 0.15in
\begin{center}
\begin{small}
\begin{sc}
\begin{tabular}{lcccr}
\toprule
Data set & Naive  & Naive+SFT & NGENT (ours) \\
\midrule
MMLU    & 68.27 & 70.83 & 69.96 \\
CMMLU & 68.00 & 71.47 & 70.37 \\
GSM8K    & 68.46 & 70.81 & 74.00 \\
IFEval    & 53.12 & 61.27 & 61.75 \\
AlignBench     & 5.26 & 5.45 & 6.50 \\
AVG      & 52.62 & 55.97 & 56.52 & \\
\bottomrule
\end{tabular}
\end{sc}
\end{small}
\end{center}
\vskip -0.1in
\end{table}

\subsection{Preliminary Results}
\paragraph{Dataset.}  
To assess the agent's personification, we conducted experiments using the CharacterEval dataset, a compilation of multi-turn role-playing dialogues specifically designed for this purpose. 
To evaluate the model's intellectual capabilities, we selected a variety of specialized datasets from the different domain, including MMLU, CMMLU, AlignBench and IFEval, which assess the knowledge and problem solving ability. Additionally, we selected GSM8K to test problem-solving skills in a mathematical context.
Each dataset offers a set of unique challenges and scenarios, rigorously testing the agent’s personification and intelligence across diverse contexts.

\paragraph{Metrics.} For the role-playing dataset, we adhere to the metrics established in CharacterEval, which primarily encompass three dimensions: conversational ability, role-playing attractiveness, and persona consistency.  For other datasets evaluating the intelligence level, we employed the accuracy metric as a standard measure of assessment.

\paragraph{Baselines.} To establish a robust comparison, we utilized a variety of models as our baselines, encompassing both general chat models and those specifically designed for role-playing. (1) For chat models, our selection included ChatGLM3-6B, XVERSE-7B/13B, Baichuan2-7B/13B, Qwen-14B, InternLM-7B/20B, and the closed-source models GPT-3.5 and GPT-4. (2) For role-playing models, we adopted CharacterGLM, Xingchen, MiniMax, and BC-NPC-Turbo, each engineered to excel in role-specific dialogues and interactions.

\paragraph{Results.} The main results are shown in Table \ref{eqresult} and Table \ref{iqresult}, which demonstrate that our method can achieve a better balance between IQ-based tasks and EQ-based tasks.

\section{Alternative Views}
While we advocate for the development of next-generation AI agents that transcend domain specialization, it is essential to acknowledge alternative viewpoints that may challenge or refine our position.

\subsection{Specialization and Domain-Specific Excellence}
One counter-argument is that specialization is the key to maximizing performance in particular areas. Many experts argue that AI systems should remain focused on specific tasks, optimizing their capabilities to an unparalleled degree within those domains. This view is supported by the success of specialized agents like DeepMind’s AlphaGo \cite{silver2017mastering}, which excels only in the specific domain of the game Go. In a similar vein, AI agents that focus on a narrow range of tasks can achieve performance levels that a general-purpose agent might struggle to match.

While specialization allows for optimal performance within a defined area, the next frontier in AI development is the integration of diverse capabilities into a single agent. A general-purpose AI need not sacrifice performance in any one domain; with advances in architecture, multi-task learning, and transfer learning, we believe that general-purpose agents can be both highly competent and adaptable. Moreover, the convergence of AI tasks—such as problem-solving, decision-making, and tool-use—will require a shift towards more integrated systems.

\subsection{Risk of Overgeneralization and Safety Concerns}
Another counter-argument is the potential risks associated with pushing AI agents to be more generalized. A highly flexible system may lack the depth of knowledge or capability required to perform specialized tasks as well as current specialized agents. Moreover, a more generalized AI could introduce ethical concerns regarding control, accountability, and safety. As AI systems become capable of handling a wider range of tasks, the complexity of ensuring that they operate in a safe and predictable manner increases.

The risks associated with general-purpose AI are valid, but they should not prevent progress. We argue that careful attention to safety mechanisms, transparency, and robust training paradigms can mitigate these risks. Furthermore, the broader potential benefits in terms of societal impact, efficiency, and accessibility justify the investment in next-generation general-purpose agents. By addressing concerns early on through research and collaboration, we can ensure that these systems evolve responsibly.

\subsection{Incremental Evolution of AI Agents}
Rather than pursuing a radical shift toward general-purpose agents, some advocate for an incremental approach where AI agents slowly evolve, with each new system combining a larger number of specialized capabilities. This method allows for more controlled development and ensures that models remain grounded in their ability to excel within specific domains. 

While an incremental approach is practical and may yield short-term benefits, we believe that the AI field would greatly benefit from a more ambitious vision of general-purpose agents. The move towards next-generation agents need not be sudden, but the long-term goal of AGI requires that we actively pursue the integration of existing specialized capabilities into unified systems. Incremental progress should support, not hinder, this vision.

\subsection{Collaboration among AI Agents} Another perspective suggests that collaboration between multiple specialized AI agents will be more effective than developing a single general-purpose AI. In this view, a multi-agent system can optimize performance by dividing tasks according to each agent's strengths, leading to more efficient and tailored solutions. Instead of forcing a single agent to excel in all areas, allowing specialized agents to work together in a coordinated manner could enable faster problem-solving without the complexity of generalization \cite{busoniu2008comprehensive,ruan2024coslight,mao2020neighborhood,mao2022seihai,chen2022ptde,xing2023dual,zhang2023controlling,ruan2024coslight}.

Collaboration between specialized agents is certainly a promising approach, and we agree that multi-agent systems have the potential to enhance performance in certain scenarios. However, the coordination and communication overhead between multiple agents could introduce inefficiencies and may hinder the system's overall adaptability in novel contexts \cite{mao2020learning}. A general-purpose agent, once it reaches a high level of competence across various domains, can potentially offer a more streamlined solution, with the ability to flexibly adjust to new situations without requiring explicit coordination between different agents. In the long term, we envision general-purpose agents as a natural extension of the collaborative frameworks. 

\section{Conclusion}
This paper advocates for the evolution of AI agents beyond their current domain-specific models, proposing that next-generation AI agents (NGENTs) should adopt a general-purpose design capable of integrating diverse functionalities across multiple domains. Such a shift is crucial not only for advancing towards true AGI but also for meeting the increasing demand for versatile, intelligent systems that can address a wide range of user needs. By focusing on seamless integration, enhanced architectural frameworks, and prioritizing safety and ethical considerations, we can create AI agents that have the potential to transform industries, solve global challenges, and improve daily life. Despite the challenges that remain, the need for this next step in AI development is pressing, and the time to push forward with this vision is now.

\bibliography{main}
\bibliographystyle{icml2025}

\end{document}